\documentclass[10pt,twocolumn,letterpaper]{article}
\usepackage[pagenumbers]{cvpr}
\usepackage{amsmath,amssymb}
\usepackage{amsthm}

\usepackage{graphicx}
\expandafter\def\csname ver@everyshi.sty\endcsname{2001/05/15}
\usepackage{tikz}
\usetikzlibrary{arrows.meta,calc,shapes.geometric}
\usepackage{pgfplots}
\pgfplotsset{compat=1.16}
\usepackage{multirow}
\makeatletter
\renewcommand{\LenToUnit}[1]{\strip@pt\dimexpr#1\relax}
\makeatother

\newcommand{\nsig}[1]{\textcolor{black!62}{#1}}
\definecolor{cvprblue}{rgb}{0.21,0.49,0.74}
\usepackage[pagebackref,breaklinks,colorlinks,allcolors=cvprblue]{hyperref}
\input{sections/numbers}
\makeatletter
\def\unsigned#1{\expandafter\unsigned@i#1\@nil}
\def\unsigned@i#1#2\@nil{\ifx+#1#2\else#1#2\fi}
\makeatother

\def\confName{CVPR}
\def\confYear{2027}
\title{What Do Learned 3D Models Add to Localization from Posed References?}
\author{%
Jaehyun Kim$^{1,*}$ \qquad Wonseok Kang$^{2,*}$ \qquad Jeongmin Lee$^{2}$ \qquad Tae-Wan Kim$^{1,\dagger}$\\[4pt]
{\normalsize $^1$Seoul National University, Department of Naval Architecture and Ocean Engineering, Seoul, South Korea}\\
{\normalsize $^2$DeepFine Co., Ltd., Seoul, South Korea}\\[2pt]
{\small\tt jaedalong@snu.ac.kr, eoid361@deepfine.com, jm.lee@deepfine.com, taewan@snu.ac.kr}\\[2pt]
{\small $^*$Equal contribution. \quad $^\dagger$Corresponding author.}%
}
\begin{document}
\twocolumn[{%
\renewcommand\twocolumn[1][]{#1}%
\maketitle
}]
\begin{abstract}
Localizing a query from a few posed reference images avoids building a 3D point map. We ask what the camera predictions
of learned 3D models add to geometric estimation from the same references, and where failures arise. Across eight learned models
and variants and five test sets, we compare reference budgets, swap pose components, and intervene on the references a
model sees. At fine pose-error thresholds, geometric estimation with one additional
reference is significantly more accurate than every model on two references on Cambridge and NAVER B1 and never
significantly worse, alone or as a fallback behind a shared acceptance gate; with the same three
references, the more accurate approach depends on the scene. The gain from the learned translation magnitude depends on reference selection: it improves seven of the eight models on 12-Scenes with the spread-out pairs we select, but none of the five
we test with the two top-ranked retrieved references. Where retrieval also returns references from other locations, at NAVER
Gangnam Station, models fail at different stages. A robust alignment brings Pi3X and MapAnything within
\rbFreeMax{} pp of adding references as needed. In controlled tests, aligning on the nearby references largely restores VGGT,
Pi3X and MapAnything but not Depth Anything 3, and pose-conditioned Pi3X recovers only if the far poses are masked and
left out of its input-pose normalization.
\end{abstract}

\begin{figure*}[t]
\centering
\includegraphics[width=\textwidth]{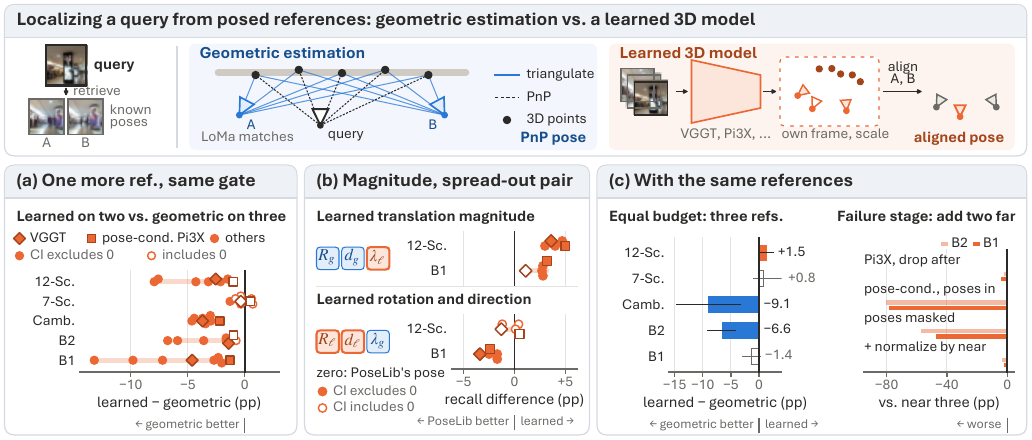}
\caption{Orange: learned, blue: geometric (subscripts $\ell$, $g$); differences are learned minus geometric; filled marks and bars: 95\% interval excludes zero; diamond: VGGT, square: pose-conditioned Pi3X, circles: other models (line: range); bars take the color of the side ahead. (a) pp over all queries (only those the shared gate rejects differ): each model on two references minus geometric estimation on three (Tab.~\ref{tab:cascade}b). (b) Recall change vs.\ PoseLib (pp) when the learned magnitude $\lambda$, or rotation $R$ and direction $d$ (magnitude refitted), replace the essential-matrix estimate's (Tab.~\ref{tab:swap}). (c) Left: pose-conditioned Pi3X on the same three references, no gate (Tab.~\ref{tab:cascade}a). Right: NAVER B2/B1, three references within 10\,m of the query plus two at least 50\,m away, minus the near three alone, aligned on the near three (Tab.~\ref{tab:intervene}).}
\label{fig:teaser}
\end{figure*}

\section{Introduction}\label{sec:intro}

Visual localization estimates a query camera pose. Many systems match image features to a prebuilt 3D point
map~\cite{hloc2019,activesearch2016}. A structureless alternative estimates the pose from retrieved reference images
with known poses and intrinsics~\cite{zhou2020learn,panek2025guide}, avoiding the need to maintain a global point map.
The remaining choices are how to estimate the query pose and when to retrieve additional references.

Geometric estimation uses query--reference correspondences, local triangulation, and PnP. Learned 3D models such as
VGGT~\cite{vggt2025}, Depth Anything 3~\cite{da3_2025}, $\pi^3$~\cite{pi3_2025}, and
MapAnything~\cite{keetha2025mapanything} predict camera poses and 3D points from sparse images, and recent localizers
combine such predictions with known reference poses~\cite{reloc3r2025,relocvggt2025,kang2026ric,l3_2026}.
Such models might reduce reliance on geometric estimation or additional references. Yet reported comparisons vary in
matcher, reference budget, and pose acceptance policy, while whole-pose accuracy obscures which learned components
help. We therefore ask what learned prediction adds beyond geometry and whether another reference is more useful.
The essential matrix determines rotation and translation direction, but not magnitude; a second posed reference
constrains magnitude, allowing it to be substituted independently. Additional references can be retrieved when the
initial pose is rejected.

\paragraph{Contributions.} We distinguish what learned camera predictions contribute from where their information is lost: reference selection changes
the gain from the learned translation magnitude, and failures arise at separable stages (\cref{fig:teaser}):

(i) \textbf{Reference budget.} Alone or behind the same shared acceptance gate, geometric estimation with three
references is significantly more accurate than every model on two on Cambridge and NAVER B1, and never significantly worse. Ungated on the same three references, pose-conditioned Pi3X is more accurate on 12-Scenes and, robustly aligned, Pi3X
on both NAVER floors and MapAnything on B1; geometric estimation stays ahead on Cambridge (\cref{sec:photos}).

(ii) \textbf{Pose components.} Substituted into the
geometric estimate, the learned translation magnitude improves 12-Scenes for seven of the eight models with our spread-out pair, for VGGT mostly on queries with few
matches supporting the geometric magnitude, but for none of the five tested with the two top-ranked retrieved references; learned
rotation and translation direction lower recall on NAVER (\cref{sec:distance}).

(iii) \textbf{Failure stage.} Keeping the retained references fixed while changing what enters inference separates failures of
the alignment after inference, of the prediction, and through the pose input. At NAVER, where retrieval also returns references
from other locations, a robust alignment after inference brings Pi3X and MapAnything within \rbFreeMax{} pp of adding references as needed
(\fvFreeStatMin--\fvFreeStatMax{} pp behind without exclusion). In controlled tests, aligning on the nearby references
largely restores VGGT, Pi3X and MapAnything but leaves Depth Anything 3 \dsDaMin--\dsDaMax{} pp below; pose-conditioned Pi3X
recovers only when the far poses are masked and left out of its input-pose normalization (\cref{sec:five}).

Where a prior comparison tests one posed-reference model against geometry~\cite{panek2025combining}, we find that
the matcher, reference budget and acceptance policy can each reverse the ranking of VGGT and geometric estimation, so we
vary one and fix the rest (\cref{sec:protocol}). Where wrongly retrieved references are removed
first~\cite{panek2025guide,han2025robustvggt}, we keep them to locate the failure: at NAVER, a robust alignment
after inference recovers most of the loss of Pi3X and MapAnything.

\section{Related Work}\label{sec:related}

\paragraph{Localization without a 3D map.} Structureless methods localize a query from posed references alone: they
intersect translation directions estimated by two-view geometry across multiple references~\cite{zhou2020learn}, or
build local models on the fly from the retrieved references~\cite{sattler2017large,panek2025guide}.
Panek~\etal~\cite{panek2025guide} survey such methods and show that they can approach methods with a 3D map. Earlier
work studies how the number of retrieved reference images affects local
triangulation~\cite{humenberger2022retrieval,pion2020retrieval}. In separate work on combining absolute and relative poses, Panek~\etal~\cite{panek2025combining} compare
structure-based and structureless methods as the database is subsampled. Given the retrieved references with their known poses, MapAnything is
less accurate, close only at the loosest tolerance; PnP on matches lifted with its depth or point maps stays far below
geometric estimation at fine tolerances, and a hybrid that can also pick a structureless pose stays below that pose alone. That comparison covers one model on one test set with 5--20 retrieved references.
Like earlier controlled comparisons in localization~\cite{sattler2017large,sattler2019limitations,brachmann2021limits}, we
control the references, test eight models and variants on five test sets, analyze individual pose components, and
compare a learned prediction with an additional reference.

\paragraph{Metric scale from learned depth.} Two-view geometry gives the rotation and the translation direction but
not its length. The feature-matching baselines of the Map-free benchmark take the length from predicted depth~\cite{arnold2022mapfree},
MicKey predicts matches in metric camera coordinates~\cite{mickey2024}, and MADPose fits the relative pose together with the scale
and shift of predicted depth maps~\cite{yu2025madpose}. Our learned-scale hybrid pose estimate (\cref{sec:setting}) takes only the length from a learned 3D model,
to isolate the effect of the learned translation magnitude.

\paragraph{Learned 3D models for localization.} DUSt3R~\cite{dust3r2024} and MASt3R~\cite{mast3r2024} predict 3D
points for pairs of reference images, aligned afterwards; VGGT~\cite{vggt2025}, Depth Anything 3~\cite{da3_2025},
$\pi^3$~\cite{pi3_2025} and MapAnything~\cite{keetha2025mapanything} predict cameras and 3D points for a few reference
images in one pass, and Depth Anything 3, MapAnything and $\pi^3$'s successor Pi3X~\cite{pi3x2025} can also take known poses as input. Localizers built
on such models regress the pose relative to retrieved reference images~\cite{reloc3r2025}, are trained to localize
from posed reference images or their features~\cite{relocvggt2025,multiloc2026,fastforward2025}, or lift learned
relative poses with the known reference poses~\cite{kang2026ric,l3_2026}. They are usually compared with other learned localizers or with pipelines that use a 3D
map, at a fixed number of reference images. We compare them with geometric estimation on the same reference images,
the geometric solvers sharing one set of matches, and vary the number of reference images.

\paragraph{Selecting views for reconstruction.}
RobustVGGT~\cite{han2025robustvggt} shows that views of other scenes degrade VGGT's reconstruction and rejects them
using its internal features. With posed references, we ask where the error enters:
in controlled tests, mostly the alignment for VGGT, Pi3X and MapAnything, the
prediction for Depth Anything 3, and the pose input for pose-conditioned Pi3X (\cref{sec:five}).

\paragraph{Reliability of learned pose estimates.} Inlier counts are the usual sign that a geometric pose is
reliable, and their image coverage adds to it~\cite{trustpose2020}. For learned 3D models, GeoCond~\cite{geocond} predicts relative-pose
reliability from conditioning cues and model-stability signals, using input-order variation as one training target. Concurrent work
examines VGGT features to identify failed relative poses~\cite{doesvggtknow}. These works assess reliability without posed
reference images. With posed references, we examine whether learned translation magnitude is reliable, how much selecting a
correct estimate would improve localization, and whether additional references benefit the learned model or geometric
estimation more.

\section{Comparison Protocol}\label{sec:setting}

\paragraph{Common experimental setup.} The database holds reference images with known poses and intrinsics; no 3D
points are stored. MegaLoc~\cite{megaloc2025} retrieves 30 reference images per query, and a D-optimal design on their
camera positions, which favors spread-out cameras, selects $K$ of them ($K$ counts the posed reference images, not the query; $K=2$ unless stated). The rule was fixed beforehand and makes two-reference geometry harder than the two top-ranked
retrieved references do (\cref{sec:distance}). Geometric estimation and every learned model receive the same selected reference images. All geometric pose estimates use
the same LoMa matches~\cite{loma2026} (\cref{sec:protocol} also uses an older matcher), and one shared
acceptance gate, based on pose uncertainty, decides which queries need another pose estimate. Additional references beyond
the initial selection are introduced only in the budget analyses of \cref{sec:protocol,sec:distance} and in \cref{sec:photos,sec:five}.
Throughout, \emph{geometric estimation} denotes pose estimation from feature correspondences, including those obtained
with a learned matcher. \emph{Learned prediction} denotes the camera estimates produced by a learned 3D model.

\paragraph{Pose components.} Let $A$ be the reference image with the most essential-matrix inliers and $B$ the other
reference image. The essential matrix between the query and $A$ gives the query rotation $R$ and a unit translation
direction $\mathbf d$ from $A$ toward the query, but not its translation magnitude $\lambda$:

\begin{equation}
  \mathbf C=\mathbf C_A+\lambda\mathbf d.
  \label{eq:split}
\end{equation}

The \emph{geometrically estimated translation magnitude} is the one-point scale supported by the raw matches to $B$. A
learned model predicts camera centers $\mathbf c$ in its own frame. The known baseline between the two reference
cameras supplies the metric scale, and the model contributes the ratio of the query--reference distance to that
baseline; their product is the \emph{learned translation magnitude}

\begin{equation}
  \lambda_{\rm learned}=s\lVert\mathbf c_q-\mathbf c_A\rVert,\qquad
  s=\frac{\lVert\mathbf C_A-\mathbf C_B\rVert}{\lVert\mathbf c_A-\mathbf c_B\rVert}.
  \label{eq:learned}
\end{equation}

The \emph{learned-scale hybrid pose estimate} keeps the geometric $(R,\mathbf d)$ and takes only $\lambda$ from the
learned model. The reverse swap takes the rotation and the translation direction together from the learned model and
fits the geometrically estimated translation magnitude again; it does not separate the rotation from the direction.
With additional reference images, the learned scale is the median over reference-image pairs.

\paragraph{Solvers and the shared acceptance gate.} Geometric estimation follows local
triangulation~\cite{panek2025guide}; here, the matches to each reference image are filtered by its essential matrix, query
keypoints shared by two reference images give tracks, and the triangulated points give 2D--3D matches for RANSAC PnP
and refinement. The position uncertainty $\sigma$ is the largest standard deviation of a Gauss--Newton covariance
scaled by the residuals, with the triangulated points treated as exact. The \emph{shared acceptance gate} accepts a pose when
$\kappa\sigma<\tau$, with $\kappa=5$ and the position tolerance $\tau$ (a larger $\kappa$ is stricter); it rejects cases with fewer than six points or
no PnP pose. $\sigma$ measures the fit, not the chance of success, so the gate can accept incorrect pose
estimates. Unless the \emph{stricter initial gate} ($\kappa=\cascKappaDev$) is used (\cref{sec:photos,sec:five}), accepted queries keep the PnP pose, so gated comparisons change only the pose estimate for the rejected queries and
measure what it adds there, not its accuracy alone; \cref{sec:protocol,sec:distance,sec:photos} also report ungated results. For rejected queries, the classical pose estimate is PoseLib's 5+1-point
generalized relative pose~\cite{poselib} with LO-RANSAC and bundle adjustment (4-pixel threshold, chosen on
development queries); \cref{tab:protocol,tab:swap} use it there.

\paragraph{Learned models.} We analyze VGGT in greatest detail. Seven more models and variants receive the same
reference images in the same order: Depth Anything 3~\cite{da3_2025}, Pi3X~\cite{pi3_2025,pi3x2025} with and without
reference-pose conditioning, MapAnything~\cite{keetha2025mapanything} with and without reference-pose conditioning,
MASt3R~\cite{mast3r2024} and Reloc3r~\cite{reloc3r2025}. Each model runs with its published weights at its own input
resolution, without task-specific training; for pose-conditioned Pi3X, we zero the masked query's pose
features, which would otherwise be NaN. Pi3X normalizes the input poses by their mean
distance from the first view, masked views included; we place the query's placeholder at the references' mean distance,
so that this scale equals the one from the references alone.
Models that accept intrinsics use the same intrinsics as geometric estimation; upright NAVER queries are padded rather
than cropped. Unless the model places the query itself, its predicted cameras are aligned to the known reference poses in closed form: rotation from the predicted against the known
reference orientations, scale from the reference baselines, then translation. VGGT's settings were chosen on development queries, the others'
are defaults.

\paragraph{Data and evaluation.} We sort 12-Scenes~\cite{twelvescenes2016} queries by scene and split alternating frames into development
(2,822) and held-out (2,819) sets. The held-out set evaluates new frames from the same rooms.
The other test sets cover different scenes: 7-Scenes~\cite{sevenscenes2013} (every 20th test frame, 850 queries),
Cambridge Landmarks~\cite{posenet2015,kendall2017geometric} (1,918), and NAVER Gangnam Station~\cite{lee2021naver}, floors B2 (916) and B1
(2,620), where the queries are the validation split and the database is the mapping split. We use structure-from-motion (SfM) reference poses~\cite{brachmann2021limits} for 7-Scenes and dataset poses for 12-Scenes. A query is correct when its position
and rotation errors are within 5\,cm and $5^\circ$ on 12-Scenes and 7-Scenes, and 25\,cm and $2^\circ$ on Cambridge
and NAVER; \emph{accuracy} is localization recall, differences are in percentage points (pp), and a query
without a pose estimate counts as incorrect. Paired 95\% intervals come from a bootstrap over scenes, or over
20-second blocks of capture time on NAVER, which has one scene per floor. Cambridge, 7-Scenes and 12-Scenes have only 5, 7 and 12 scenes to resample, so
their intervals may be too narrow.

\paragraph{Evaluation subsets.} The swaps need an essential matrix with $A$ and the model's output; these exist for all
but 0--7 queries per test set, except NAVER B1, where 328 queries have no essential matrix. \Cref{tab:swap} uses the
queries where the swapped estimate exists (a missing PoseLib pose counts as incorrect); \cref{sec:photos} counts every
query. Solver thresholds, translation-magnitude priors and the settings for adding references after gate rejection (\cref{sec:photos}) are selected on
development queries. The other classical controls and the reference-subset analysis (\cref{sec:five}) followed,
with all settings fixed; the subset analysis is exploratory. Reference images fetched are counted per query; running times
are measured on one machine.

\section{Sensitivity to the Comparison Protocol}\label{sec:protocol}

Before analyzing individual pose components, we examine how the comparison protocol changes the relative performance
of VGGT and geometric estimation. \Cref{tab:protocol} uses VGGT's query pose aligned to the same known reference poses
and varies one experimental choice at a time.

\begin{table}[t]
\centering\footnotesize\setlength{\tabcolsep}{3.2pt}
\caption{Sensitivity to the comparison protocol: gain in recall (pp) of VGGT's aligned pose over geometric estimation (positive: VGGT ahead), changing one choice per block; read down a column. \emph{Alone}: against PnP, every query, no gate. \emph{Gate}: both keep the PnP pose the shared gate accepts; on rejected queries, VGGT against PnP or PoseLib. B2, B1: NAVER floors; the 886 and 2,544 queries with stored SuperPoint+LightGlue matches. Pose-cond.\ Pi3X: its own pose instead of VGGT's, same queries. Gray: 95\% interval includes zero.}
\label{tab:protocol}
\begin{tabular}{@{}lccccc@{}}
\toprule
 & 12-Sc. & 7-Sc. & Camb. & B2 & B1 \\
\midrule
\multicolumn{6}{@{}l}{\emph{SuperPoint+LightGlue, two references}} \\
\quad alone & +25.4 & +7.3 & $-$26.6 & +15.8 & +33.3 \\
\quad gate, PnP where rejected & +25.2 & +8.6 & $-$1.9 & +17.5 & +32.8 \\
\quad gate, PoseLib where rejected & +12.7 & \nsig{+3.4} & $-$2.3 & +5.2 & +14.0 \\
\multicolumn{6}{@{}l}{\emph{LoMa, two references}} \\
\quad alone & +9.0 & $-$1.5 & $-$27.3 & $-$5.8 & \nsig{$-$0.7} \\
\quad pose-cond.\ Pi3X, alone & +12.5 & \nsig{+1.4} & \nsig{$-$2.5} & +4.6 & +9.2 \\
\quad gate, PnP where rejected & +8.5 & +0.9 & \nsig{$-$1.1} & \nsig{+1.7} & +3.7 \\
\quad gate, PoseLib where rejected & +2.4 & \nsig{$-$0.1} & $-$1.4 & \nsig{$-$0.9} & \nsig{$-$0.2} \\
\multicolumn{6}{@{}l}{\emph{LoMa, four references}} \\
\quad alone & $-$5.6 & $-$0.9 & $-$30.8 & $-$12.1 & $-$11.8 \\
\quad gate, PnP where rejected & \nsig{+0.1} & \nsig{+0.5} & \nsig{$-$0.1} & \nsig{$-$0.1} & \nsig{$-$0.3} \\
\quad gate, PoseLib where rejected & $-$0.5 & \nsig{+0.2} & \nsig{$-$0.5} & \nsig{$-$0.6} & $-$2.2 \\
\bottomrule
\end{tabular}

\end{table}

\paragraph{The matcher.} With SuperPoint and
LightGlue~\cite{superpoint2018,lightglue2023} and two references, VGGT, evaluated alone, leads geometric
estimation on four of five test sets, by up to \unsigned{\prSplgTwoNavOneAlone} pp. With LoMa on the same references and evaluated
alone, VGGT retains an advantage only on 12-Scenes; on the other sets, it shows no statistically significant
gain or performs worse.

\paragraph{The number of reference images.} Increasing the budget to four references with LoMa favors geometric
estimation: evaluated alone, VGGT is less accurate on all five test sets, and behind the shared acceptance
gate it is not significantly better on any. The four-reference selection contains the original pair for 98--100\% of
queries, except NAVER B2 (91\%), where few of the retrieved references depict the most likely location and the
selection can draw from all 30.

\paragraph{Evaluation policy.} When evaluated
alone, VGGT is \unsigned{\prLomaTwoTwHoAlone} pp ahead of PnP on 12-Scenes but \prLomaTwoCambAloneAbs{} pp
behind on Cambridge. A shared acceptance gate that retains accepted PnP estimates reduces the Cambridge difference to
\prLomaTwoCambPk{} pp, while leaving the 12-Scenes difference at \prLomaTwoTwHoPk{} pp. Replacing PnP for rejected
queries with PoseLib's generalized relative pose further reduces the 12-Scenes difference to \prLomaTwoTwHoChk{} pp; on NAVER
B1, the difference changes from \prLomaTwoNavOnePk{} to \prLomaTwoNavOneChk{} pp. Thus the gate and fallback determine
which queries contribute to the measured gain.

Overall, an older matcher and fewer references favor VGGT's predictions; pose-conditioned Pi3X, evaluated alone,
still outperforms PnP with LoMa and two references on 12-Scenes and both NAVER floors (\cref{tab:protocol}).
Evaluation alone and the shared acceptance gate measure
different quantities: behind the gate both keep the same accepted PnP pose, so the difference counts only what each estimate
adds for rejected queries, and with a classical estimate there all fifteen differences in \cref{tab:protocol} are closer to zero than alone. The remaining comparisons therefore fix LoMa, specify the
reference budget, and apply one shared acceptance gate.

\section{Learned Pose Components with Two References}\label{sec:distance}

For queries rejected by the shared acceptance gate, we replace either the translation magnitude or the rotation and
translation direction of a geometric estimate with the corresponding learned predictions and compare the result with
PoseLib's pose. To separate the parts, the geometric estimate is our essential-matrix decomposition:
rotation and translation direction from the essential matrix of reference image $A$, one-point translation magnitude
from reference image $B$; alone, it is \swDecompCamb{} to \swDecompNavOne{} pp from PoseLib, significantly only on B1, and against the decomposition itself, where it exists, the same 12-Scenes, B2 and B1 gains are significant (\cref{tab:swap}). The learned parts
come from the model's cameras aligned to the two reference images, with the pose estimate placed at reference image $A$. For the other three models, the learned magnitude also improves 12-Scenes for MASt3R and
pose-conditioned MapAnything (\swXDistMtTwHo{} and \swXDistMaPcTwHo{} pp), not for Reloc3r.

\begin{table}[t]
\centering\footnotesize\setlength{\tabcolsep}{2.5pt}
\caption{Pose components with two references, for queries rejected by the shared acceptance gate: change in recall (pp over all paired queries; positive: learned part ahead) over PoseLib's pose when the model supplies only the translation magnitude, or only the rotation and translation direction, of our essential-matrix decomposition. \emph{Decomposition alone}: no learned part, also against PoseLib. The learned magnitude helps most on 12-Scenes; learned rotation and direction lower recall on NAVER. Gray: 95\% interval includes zero.}
\label{tab:swap}
\begin{tabular}{@{}lccccc@{}}
\toprule
\emph{$\Delta$ recall vs.\ PoseLib (pp)} & 12-Sc. & 7-Sc. & Camb. & B2 & B1 \\
\midrule
\multicolumn{6}{@{}l}{\emph{learned translation magnitude, geometric rotation and direction}} \\
VGGT & +3.6 & \nsig{+0.4} & $-$0.8 & +1.5 & \nsig{+1.1} \\
Depth Anything 3 & +4.7 & \nsig{+0.4} & $-$0.9 & +2.1 & +2.8 \\
Pi3X & +4.0 & \nsig{+0.4} & \nsig{$-$0.3} & \nsig{+1.0} & +2.7 \\
Pi3X, pose-cond. & +5.0 & \nsig{+0.5} & \nsig{+0.3} & +1.3 & +3.2 \\
MapAnything & +3.0 & \nsig{$-$0.1} & \nsig{$-$0.6} & \nsig{+1.0} & +2.8 \\
\midrule
\multicolumn{6}{@{}l}{\emph{learned rotation and direction, geometric translation magnitude}} \\
VGGT & \nsig{$-$1.3} & \nsig{$-$0.4} & $-$1.5 & $-$2.8 & $-$3.4 \\
Depth Anything 3 & \nsig{+0.4} & +0.4 & $-$1.4 & $-$2.3 & $-$2.5 \\
Pi3X & \nsig{+0.1} & +0.4 & \nsig{$-$0.5} & $-$2.2 & $-$1.7 \\
Pi3X, pose-cond. & \nsig{+0.5} & +0.4 & \nsig{$-$0.7} & $-$1.6 & $-$2.4 \\
MapAnything & \nsig{$-$1.2} & \nsig{$-$0.1} & \nsig{$-$0.5} & $-$1.7 & $-$1.7 \\
\midrule
decomposition alone & \nsig{$-$0.3} & \nsig{$-$0.4} & \nsig{$-$0.6} & \nsig{+0.1} & +1.0 \\
\midrule
paired queries (VGGT) & 2,814 & 850 & 1,918 & 909 & 2,292 \\
\bottomrule
\end{tabular}

\end{table}

\paragraph{Pose-component contributions.} With our two references, translation magnitude is the clearest useful learned
pose component. Substituting the learned magnitude improves held-out 12-Scenes for all five models
by \swDistTwHoMin{} to \swDistTwHoMax{} pp (\dcDistTwHoMin{} to \dcDistTwHoMax{} against the decomposition where it exists); it also improves NAVER B1 for every model except VGGT, and NAVER B2 for
VGGT, Depth Anything 3, and pose-conditioned Pi3X. It yields no significant gain on 7-Scenes or Cambridge. By contrast, the reverse swap, with the learned rotation and translation
direction and a refitted geometric magnitude, does not significantly change 12-Scenes, gains at most 0.4 pp on 7-Scenes, and reduces accuracy on both NAVER
floors for every model by \swRotNavMin{} to \swRotNavMax{} pp.
The effects need not be additive: on B1, all components of
pose-conditioned Pi3X gain \swAllPipNavOne{} pp, more than its magnitude alone (\swDistPipNavOne), although its rotation and
direction alone give \swRotPipNavOne; for VGGT, all components together score below its magnitude alone.

\paragraph{Reference selection and budget.} The gain depends on the selected pair. With the two top-ranked
references in retrieval order instead of the D-optimal pair (a rule fixed beforehand), the gate rejects \rsRejHoR\% of
12-Scenes queries instead of 18\%, and the learned magnitude shows no significant gain on 12-Scenes, B2 or B1 for any of the
five swap models. The difference persists without the gate: against the decomposition with the same rotation and direction, VGGT's magnitude
changes 12-Scenes recall by \wgAllDHo{} pp [\wgAllDHoLo, \wgAllDHoHi] with the D-optimal pair and by \wgAllTHo{}
[\wgAllTHoLo, \wgAllTHoHi] with the top-ranked pair. We call \emph{support} the number of matches of reference $B$ that
agree with the geometric magnitude, by the inlier test of its fit, and \emph{low support} the lowest third of this count over
both selections together (below \wgCutLoHo{} matches), a cutoff set without the true pose. Low support covers \wgLowShareDHo\% of the
D-optimal queries, which gain \wgLowDHo{} pp [\wgLowDHoLo, \wgLowDHoHi], but only \wgLowShareTHo\% of the top-ranked queries
(\wgLowTHo{} [\wgLowTHoLo, \wgLowTHoHi]; difference \unsigned{\wgDiffHo} pp [\wgDiffHoLo, \wgDiffHoHi]). The two groups hold different queries, so this shows where the gain
lies, not an effect of support alone.
With four references, VGGT's learned magnitude yields no significant improvement on any test set.

\paragraph{Comparison with other classical pose solvers.} On held-out 12-Scenes, VGGT's learned-scale hybrid estimate
(92.5\% behind the gate) exceeds an oracle choosing the better of PoseLib and our refined decomposition with the true pose
(90.1\%). It stays \srAllGain{} pp ahead (95\% CI [\srAllLo, \srAllHi]) when PnP on raw matches and on points triangulated from
the two references is tried first, and constant and baseline-proportional magnitudes fitted on development queries reach
only 80--82\%.

\paragraph{Selecting between the magnitudes.} On held-out 12-Scenes, even an oracle choice between the two magnitudes adds only 0.7 pp over always using the
learned-scale hybrid estimate. We next ask whether an additional reference is better used to compute a new pose.

\section{One Additional Reference versus a Learned Model}\label{sec:photos}

We ask whether one more reference for geometric estimation is more useful than running a model on the two
available.

\paragraph{Equal budgets.} We add the next retrieved reference to the original pair and give the same three references
to geometric estimation (local triangulation and PnP for every query) and to pose-conditioned Pi3X, both without a gate
(\cref{tab:cascade}a). Pi3X is \eqAloneTwHo{} pp more accurate on held-out 12-Scenes and does not differ significantly
on 7-Scenes and NAVER B1, whereas geometric estimation is \eqAloneCamb{} pp more accurate on Cambridge and
\eqAloneNavTwo{} pp on NAVER B2. With the robust alignment of \cref{sec:five}, Pi3X without pose input is more accurate
on both NAVER floors and MapAnything on B1 (\eqrNavMin--\eqrNavMax{} pp), and geometric estimation stays ahead on Cambridge (\cref{tab:cascade}a).

\paragraph{One more reference against a learned fallback.} With the same shared acceptance gate on both sides (accepted
queries keep the two-reference PnP pose), geometric estimation on three references for the rejected queries is significantly
more accurate than every two-reference model on Cambridge (\sgCambMin--\sgCambMax{} pp) and NAVER B1 (\sgNavOneMin--\sgNavOneMax{}
pp), than seven of the eight on held-out 12-Scenes (pose-conditioned Pi3X trails by \sgTwHoPip{} pp, not significantly) and than five on B2; on
7-Scenes only the difference to MASt3R is significant, and no model is significantly ahead on any test set
(\cref{tab:cascade}b). Using the third reference for every query also replaces accepted
two-reference poses, which adds \sgCambAcc{} pp on Cambridge and \sgTwHoAcc--\sgNavOneAcc{} pp on 12-Scenes and B1, which no
fallback can give.

The lead of ungated three-reference estimation over gated two-reference learned estimates depends on precision. It narrows at looser thresholds,
is no longer clear on 12-Scenes at 10\,cm, $5^\circ$, and disappears or reverses on NAVER at 1\,m, $5^\circ$ and 2\,m, $10^\circ$. Against VGGT, it remains when references are taken in retrieval
order rather than selected geometrically. Matching the added reference takes \tmStepMtOne\,ms and solving the pose
\tmStepCumOne\,ms on held-out 12-Scenes, against \tmStepVggtThree\,ms for one VGGT pass.

\begin{table}[t]
\centering\footnotesize\setlength{\tabcolsep}{1.9pt}
\caption{Additional references. (a) Recall without a gate, the same references for both: geometric estimation on three; pose-conditioned Pi3X on the same three or the first two, and Pi3X and MapAnything on the same three with the robust alignment of Tab.~\ref{tab:five} (gray: not significantly different from geometric estimation). (b) Where the shared gate rejects, each model on two references minus geometric estimation on three, both keeping the accepted PnP pose (pp over all queries; gray: 95\% interval includes zero); last rows: the accepted queries and the rejected share. Model estimate: VGGT's learned-scale hybrid, pose-conditioned MapAnything's own pose, otherwise the better of the two per test set (a choice that favors the models). (c) VGGT's hybrid estimate as soon as the gate rejects, against add refs.}
\label{tab:cascade}
\begin{tabular}{@{}lccccc@{}}
\toprule
 & 12-Sc. & 7-Sc. & Camb. & B2 & B1 \\
\midrule
\multicolumn{6}{@{}l}{\emph{(a) the same references, no gate (recall, \%)}} \\
geometric estimation, three & 95.9 & 96.9 & 61.7 & 50.1 & 58.3 \\
pose-cond.\ Pi3X, the same three & 97.4 & \nsig{97.8} & 52.6 & 43.6 & \nsig{56.9} \\
pose-cond.\ Pi3X, two & \nsig{95.1} & \nsig{96.1} & 48.5 & \nsig{48.6} & 56.1 \\
Pi3X, the same three, robust & \nsig{96.2} & \nsig{97.1} & 52.1 & 52.2 & 60.2 \\
MapAnything, the same three, robust & \nsig{95.1} & 91.3 & 46.5 & \nsig{49.2} & 60.6 \\
\midrule
\multicolumn{6}{@{}l}{\emph{(b) one more reference against each model, same gate (pp, all queries)}} \\
\multicolumn{6}{@{}l}{\quad only rejected queries differ: model on two minus geometric on three} \\
\quad VGGT & $-$2.6 & \nsig{$-$0.4} & $-$3.7 & $-$1.4 & $-$4.6 \\
\quad Depth Anything 3 & $-$1.5 & \nsig{+0.6} & $-$3.7 & \nsig{$-$0.8} & $-$2.7 \\
\quad Pi3X & $-$2.1 & \nsig{+0.7} & $-$3.0 & \nsig{$-$1.6} & $-$2.0 \\
\quad Pi3X, pose-cond. & \nsig{$-$1.0} & \nsig{+0.5} & $-$2.2 & \nsig{$-$1.0} & $-$1.3 \\
\quad MapAnything & $-$3.2 & \nsig{$-$0.4} & $-$3.4 & $-$1.6 & $-$2.3 \\
\quad MapAnything, pose-cond. & $-$4.3 & \nsig{$-$0.4} & $-$3.0 & $-$3.6 & $-$7.2 \\
accepted queries: two minus three & $-$0.9 & \nsig{$-$0.7} & $-$7.9 & \nsig{$-$1.2} & $-$1.9 \\
queries the gate rejects (\%) & 17.8 & 3.6 & 23.0 & 36.4 & 47.7 \\
\midrule
\multicolumn{6}{@{}l}{\emph{(c) more references only where a gate rejects (recall, \%)}} \\
two references, then VGGT's hybrid & 92.5 & 95.9 & 50.1 & 47.5 & 51.8 \\
add refs & 97.4 & 97.3 & 60.9 & 59.0 & 65.6 \\
\bottomrule
\end{tabular}

\end{table}

\paragraph{Adding references after gate rejection.} The procedure we evaluate (add refs) adds references as needed before using a learned fallback. It keeps the
two-reference pose only when the stricter initial gate ($\kappa=\cascKappaDev$, set on development queries) accepts it.
Otherwise, it adds the next retrieved references one at a time, keeping them, until the shared acceptance gate accepts a
pose or five references are in use (\cref{tab:cascade}c). If none is accepted, it keeps the two-reference pose when the
shared acceptance gate accepted it, and otherwise returns VGGT's learned-scale hybrid estimate. The comparator uses that hybrid
estimate as soon as the shared acceptance gate rejects.

Re-checking the two-reference pose with the stricter gate and adding references improves recall over this comparator by \unsigned{\cascTwHoCumGain} to \unsigned{\cascNavOneCumGain} pp
on held-out 12-Scenes, Cambridge, and both NAVER floors, and slightly on 7-Scenes (\cref{tab:cascade}c; differences
taken before rounding). It uses \nlPhotosLo--\nlPhotosHi{} additional references per query on average, depending on
the test set. The stricter initial gate matters: the shared acceptance gate accepts incorrect two-reference poses
for \diagCambWrongAcc\% of all Cambridge queries and \diagNavTwoWrongAcc\% of NAVER B2 queries.

\paragraph{Incorrect estimates and fallback behavior.} Accepted poses can be incorrect: the gate tests pose fit, and
this procedure still returns an incorrect pose for \waCascadeCambWrong\% of Cambridge queries and \waCascadeNavTwoWrong\% of NAVER B2 queries. Triangulating from all five
references for every query is 0.8--7.1 pp more accurate (\cref{tab:five}) than adding references as
needed, but uses three additional references for every query instead of at
most \nlPhotosHi{} on average.

\section{Alignment and Pose-Input Failures}\label{sec:five}

Here, each of the eight learned models and variants receives the
same pool of five references---the initial two and the three that may be retrieved after gate rejection---under the same stricter
initial gate (\cref{tab:five}). The procedure adds references one at a time and can stop early, whereas each learned model processes all five at once. Replacing the procedure's
learned fallback with a classical estimate changes recall by no more than 0.4 pp, so the comparison
primarily reflects geometric estimation.

\begin{table}[t]
\centering\footnotesize\setlength{\tabcolsep}{2.2pt}
\caption{The same five references for every model: its own pose where the stricter initial gate rejects, minus adding references as needed (add refs; pp over all queries; negative: add refs ahead; a missing estimate counts as incorrect). Model rows: its own pose aligned over all five (mean rotation and center offset); pose-conditioned MapAnything and Reloc3r return map poses themselves. Indented: inconsistent references excluded after inference (a robust alignment of the same predictions) or, for pose-conditioned Pi3X, before it (its most compact three of the five). Geometric estimation, all five: triangulation and PnP on all five for every query. Gray: 95\% interval includes zero.}
\label{tab:five}
\begin{tabular}{@{}lccccc@{}}
\toprule
 & 12-Sc. & 7-Sc. & Camb. & B2 & B1 \\
\midrule
add refs (recall, \%) & 97.4 & 97.3 & 60.9 & 59.0 & 65.6 \\
geometric estimation, all five & +1.0 & +0.8 & +7.1 & +3.4 & +1.8 \\
\midrule
VGGT & $-$1.7 & \nsig{+0.2} & $-$13.3 & $-$16.7 & $-$18.9 \\
\quad robust alignment, after & \nsig{$-$0.4} & \nsig{+0.6} & $-$13.4 & $-$6.9 & $-$4.8 \\
Depth Anything 3 & $-$2.0 & \nsig{+0.7} & $-$10.6 & $-$15.9 & $-$18.2 \\
\quad robust alignment, after & \nsig{0.0} & \nsig{+1.1} & $-$10.8 & $-$6.3 & $-$8.6 \\
Pi3X & $-$1.8 & +1.9 & $-$4.5 & $-$15.7 & $-$16.8 \\
\quad robust alignment, after & \nsig{$-$0.2} & +1.9 & \nsig{$-$4.0} & \nsig{$-$2.0} & \nsig{$-$1.5} \\
Pi3X, pose-cond. & \nsig{+0.1} & +1.8 & $-$3.4 & $-$15.0 & $-$14.4 \\
\quad robust alignment, after & \nsig{+0.1} & +1.6 & $-$3.3 & $-$14.5 & $-$13.8 \\
\quad most compact three, before & \nsig{$-$0.1} & $-$4.1 & $-$11.3 & $-$2.5 & $-$2.5 \\
MapAnything & $-$3.6 & \nsig{$-$2.7} & $-$6.5 & $-$15.9 & $-$16.8 \\
\quad robust alignment, after & \nsig{$-$0.6} & \nsig{$-$2.6} & $-$6.2 & $-$3.3 & \nsig{0.0} \\
MapAnything, pose-cond. & $-$4.9 & $-$5.6 & $-$13.7 & $-$15.4 & $-$20.8 \\
MASt3R & $-$11.3 & $-$10.0 & $-$17.8 & $-$21.2 & $-$31.9 \\
\quad robust alignment, after & $-$9.2 & $-$9.9 & $-$17.9 & $-$18.3 & $-$28.1 \\
Reloc3r & $-$4.7 & \nsig{$-$3.3} & $-$16.9 & $-$22.6 & $-$38.1 \\
\bottomrule
\end{tabular}

\end{table}

\providecolor{inkgray}{HTML}{52514E}\providecolor{gridgray}{HTML}{E4E3DF}\providecolor{newblue}{HTML}{2A78D6}\providecolor{vggtorange}{HTML}{EB6834}\providecolor{robteal}{HTML}{1B9E77}
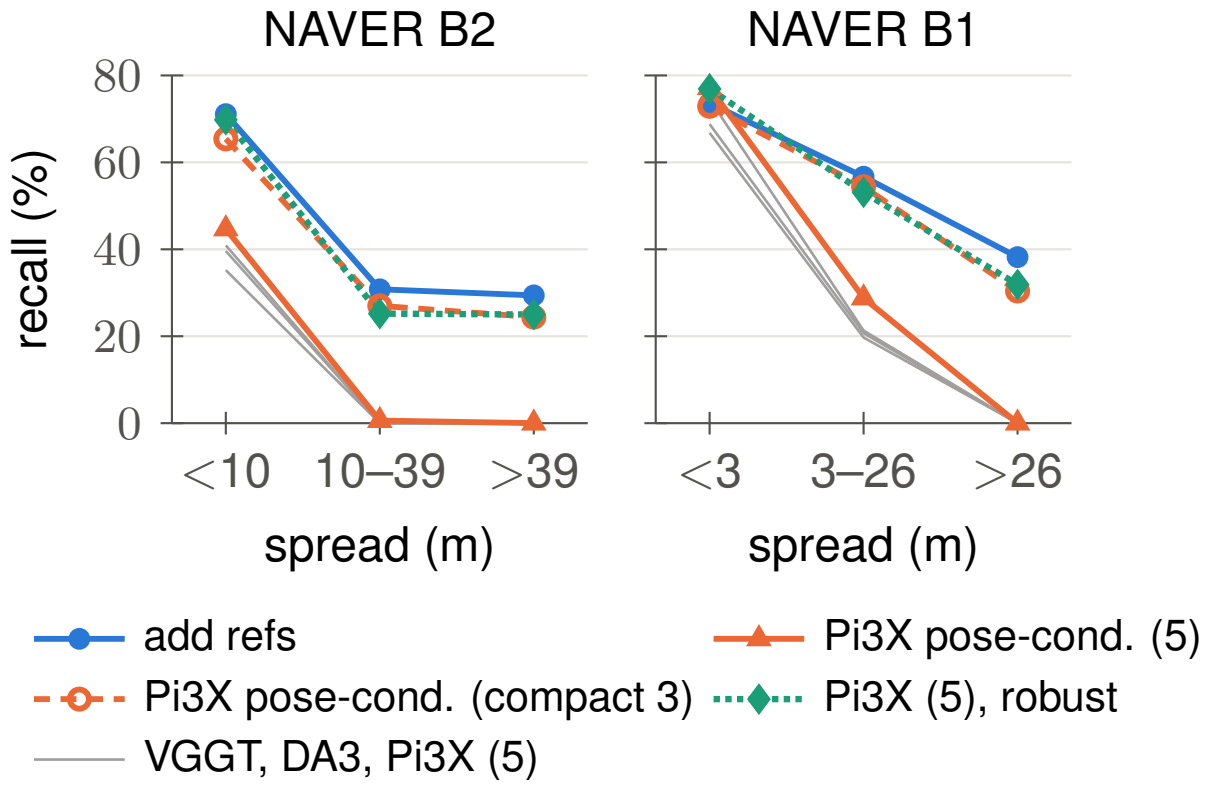
\begin{figure}[t]
\centering
\begin{tikzpicture}
\begin{axis}[at={(0.00cm,0)}, width=2.75cm, height=2.3cm, scale only axis=true, clip=false, xmin=-0.35, xmax=2.35, xtick={0,1,2}, xticklabels={$<$10,10--39,$>$39}, ymin=0, ymax=80, ytick={0,20,40,60,80}, axis lines=left, axis line style={inkgray, line width=0.4pt, -}, tick style={inkgray, line width=0.4pt}, tick label style={font=\small\sffamily, inkgray}, label style={font=\small\sffamily}, ymajorgrids, major grid style={gridgray, line width=0.4pt}, title style={font=\small\sffamily, yshift=-4pt}, title={NAVER B2}, ylabel={recall (\%)}, legend cell align=left, legend style={at={(1.082,-0.5)}, anchor=north, legend columns=2, font=\footnotesize\sffamily, draw=none, column sep=1pt}, xlabel={spread (m)}]
\addplot[inkgray!55, solid, line width=0.5pt, mark=none, forget plot] coordinates {(0,35.2) (1,0.0) (2,0.0)};
\addplot[inkgray!55, solid, line width=0.5pt, mark=none, forget plot] coordinates {(0,39.6) (1,0.0) (2,0.0)};
\addplot[inkgray!55, solid, line width=0.5pt, mark=none, forget plot] coordinates {(0,40.9) (1,0.0) (2,0.0)};
\addplot[newblue, solid, line width=1.1pt, mark=*, mark size=1.5pt] coordinates {(0,71.1) (1,30.8) (2,29.4)};
\addplot[vggtorange, solid, line width=1.1pt, mark=triangle*, mark size=1.7pt] coordinates {(0,44.7) (1,0.6) (2,0.0)};
\addplot[vggtorange, densely dashed, line width=1.1pt, mark=o, mark size=1.8pt, mark options={solid, fill=white}] coordinates {(0,65.4) (1,27.0) (2,24.4)};
\addplot[robteal, densely dotted, line width=1.2pt, mark=diamond*, mark size=2.0pt, mark options={solid}] coordinates {(0,69.8) (1,25.2) (2,25.0)};
\addlegendimage{inkgray!55, solid, line width=0.5pt, mark=none}
\legend{{add refs},{Pi3X pose-cond. (5)},{Pi3X pose-cond. (compact 3)},{Pi3X (5), robust},{VGGT, DA3, Pi3X (5)}}
\end{axis}
\begin{axis}[at={(3.20cm,0)}, width=2.75cm, height=2.3cm, scale only axis=true, clip=false, xmin=-0.35, xmax=2.35, xtick={0,1,2}, xticklabels={$<$3,3--26,$>$26}, ymin=0, ymax=80, ytick={0,20,40,60,80}, axis lines=left, axis line style={inkgray, line width=0.4pt, -}, tick style={inkgray, line width=0.4pt}, tick label style={font=\small\sffamily, inkgray}, label style={font=\small\sffamily}, ymajorgrids, major grid style={gridgray, line width=0.4pt}, title style={font=\small\sffamily, yshift=-4pt}, title={NAVER B1}, yticklabels={}, xlabel={spread (m)}]
\addplot[inkgray!55, solid, line width=0.5pt, mark=none, forget plot] coordinates {(0,66.8) (1,19.7) (2,0.0)};
\addplot[inkgray!55, solid, line width=0.5pt, mark=none, forget plot] coordinates {(0,68.8) (1,20.7) (2,0.0)};
\addplot[inkgray!55, solid, line width=0.5pt, mark=none, forget plot] coordinates {(0,74.5) (1,21.3) (2,0.0)};
\addplot[newblue, solid, line width=1.1pt, mark=*, mark size=1.5pt] coordinates {(0,73.6) (1,56.7) (2,38.2)};
\addplot[vggtorange, solid, line width=1.1pt, mark=triangle*, mark size=1.7pt] coordinates {(0,77.2) (1,28.8) (2,0.0)};
\addplot[vggtorange, densely dashed, line width=1.1pt, mark=o, mark size=1.8pt, mark options={solid, fill=white}] coordinates {(0,72.9) (1,54.4) (2,30.4)};
\addplot[robteal, densely dotted, line width=1.2pt, mark=diamond*, mark size=2.0pt, mark options={solid}] coordinates {(0,76.9) (1,53.1) (2,31.9)};
\end{axis}
\end{tikzpicture}
\caption{NAVER queries rejected by the stricter initial gate, by thirds of the spread of the five reference cameras around their centroid; (5): all five references (MapAnything, Reloc3r, MASt3R, not shown: at most 1\% in the widest third). Queries: 478 (B2) and 1,815 (B1).}
\label{fig:spread}
\end{figure}

\begin{table}[t]
\centering\footnotesize\setlength{\tabcolsep}{3pt}
\begin{tabular}{@{}lcccccc@{}}
\toprule
 & far & norm. & \multicolumn{2}{c}{B2 (636)} & \multicolumn{2}{c}{B1 (1802)} \\
\cmidrule(lr){4-5}\cmidrule(l){6-7}
 & poses & over & near & +far & near & +far \\
\midrule
geometric estimation & -- & -- & 85.7 & 84.7 & 78.6 & 78.5 \\
Pi3X & -- & -- & 88.1 & 86.2 & 80.9 & 76.7 \\
VGGT & -- & -- & 73.1 & 72.3 & 73.6 & 72.1 \\
Depth Anything 3 & -- & -- & 83.2 & 55.5 & 74.4 & 39.3 \\
MapAnything & -- & -- & 83.8 & 82.5 & 79.7 & 79.3 \\
pose-cond.\ MapAnything & given & -- & 75.2 & 1.6 & 67.6 & 1.9 \\
pose-cond.\ Pi3X & given & all & 89.2 & 8.5 & 83.7 & 5.2 \\
 & given & near &  & 0.0 &  & 0.9 \\
 & masked & all &  & 32.1 &  & 36.1 \\
 & masked & near &  & 85.8 &  & 81.8 \\
\bottomrule
\end{tabular}
\caption{Failure stages at NAVER (recall, \%). Near: three references within 10\,m of the query; +far: two more at least 50\,m away (chosen by true position only). Models are aligned on the near three after inference (pose-conditioned MapAnything: its own map poses); geometric estimation uses three or five references. Norm.\ over: views whose mean distance normalizes input poses. Aligned over all five, models without pose input reach 0\%.}
\label{tab:intervene}
\end{table}

\paragraph{Room-scale and Cambridge results.} With the same five-reference pool, pose-conditioned Pi3X is not
significantly different from add refs at the 95\% level on held-out 12-Scenes. Pi3X exceeds
it on 7-Scenes with and without reference-pose conditioning. On Cambridge, without exclusion, every learned model is lower at 25\,cm (\fvCambPiPosed{} to
\fvCambOthMax{} pp); at 1\,m, $5^\circ$, pose-conditioned Pi3X leads.

\paragraph{NAVER results and reference spread.} NAVER gives a different result: with all five references and no
exclusion, every learned model is below add refs by \fvStatMin{} to \fvStatMax{} pp. For queries rejected by the stricter initial gate, the median spread around the five camera centers'
centroid is \fvSpreadNavTwo\,m on NAVER B2 and \fvSpreadNavOne\,m on B1 (\fvSpreadTwHo\,m on 12-Scenes). For the most dispersed third (\cref{fig:spread}), all eight learned models without reference exclusion localize only
0--1\% of queries, versus \fvCascNavOneWide\% for add refs on B1 and \fvCascNavTwoWide\% on B2.

\paragraph{References from other locations.} Widely spread references usually include some from other locations, which the
models place inconsistently: after the alignment over all five references, Depth
Anything 3 and Pi3X without pose conditioning leave errors of \fvResNavTwoFree\% of reference spread on B2 and \fvResNavOneFree\% on B1. A robust alignment changes only the references used to estimate the alignment, not the predicted cameras, and needs no true pose: it aligns to each pair of references, keeps the pair most other
references agree with (within a quarter of the median reference distance and $10^\circ$, the middle of three settings), and
refits on those. Pi3X and
MapAnything without pose input are then within \rbFreeMax{} pp of add refs on both floors, and VGGT and Depth
Anything 3 are lower by \rbVdMin--\rbVdMax{} pp (\cref{tab:five}), and Pi3X keeps part of its recall in the most dispersed third (\cref{fig:spread}).
Pose-conditioned Pi3X changes little (\rbPipNavTwo{} and \rbPipNavOne{} pp
lower); MASt3R remains \rbMtMin--\rbMtMax{} pp lower.

\paragraph{Spatially compact reference subsets.}
Among the ten three-reference subsets of the five, the most compact one (smallest mean distance to its centroid, from
the known reference positions) narrows pose-conditioned Pi3X's gap to add refs from about 15 to 2.5 pp on both floors
(\cref{tab:five}). Where all five show one place, the
compact three trail add refs by 4.1 and 11.3 pp on 7-Scenes and Cambridge.
The most compact three exceed the most dispersed by \sbCWNavTwo{} and \sbCWNavOne{} pp on B2 and B1, \unsigned{\sgDiDNavTwo} and
\unsigned{\sgDiDNavOne} pp more than for geometric estimation (intervals above zero).
A controlled test keeps the retained references fixed and changes only
what enters inference (\cref{tab:intervene}): adding two far references and aligning on the near three after inference
restores Pi3X, VGGT and MapAnything to within \dsFreeAfterMax{} pp of the three alone but leaves Depth Anything 3 \dsDaMin--\dsDaMax{} pp
below, so the far images also degrade its prediction, and pose-conditioned Pi3X \dsPosedAfterMin--\dsPosedAfterMax{} pp below; given the
far poses, pose-conditioned MapAnything also drops by \dsMaPcMin--\dsMaPcMax{} pp. Masking the far poses restores pose-conditioned Pi3X
only if the near three alone also normalize its input poses. At Cambridge, given the far poses, pose-conditioned Pi3X drops by \dcbPosed{} pp.

\paragraph{Implications.} At NAVER, most of the deficit of most models thus comes from references at other places, and in our controlled tests
the failure stages need different handling. For VGGT, Pi3X and MapAnything, use alignment robust to inconsistent references;
Depth Anything 3 also needs them left out before inference. For pose-conditioned Pi3X as evaluated, a reference whose pose is masked must also be left out of the
normalization of its input poses. The robust alignment and the compact three choose references without the true pose;
the controlled tests use it only to compose the sets, not as a selection rule.

\section{Limitations}\label{sec:limits}

Our conclusions are limited to the tested solvers, matcher, alignment, reference selection rule, and the camera predictions of eight
learned models and variants, evaluated with published weights and the inference adaptations of \cref{sec:setting},
without task-specific training.
Pose-conditioned Pi3X's training data
are not published, whereas the reported training data for $\pi^3$ exclude our test sets. The geometry-based pipeline
uses a learned matcher, so removing the learned 3D model does not remove all learning. The held-out 12-Scenes queries
come from the same rooms as the development queries, and the NAVER result comes from two floors of one station, with a controlled check at Cambridge. Our oracle ceilings cover only selection among the
estimates we computed. Analyses that use the true pose explain failures and are not rules, and differences whose intervals include zero do
not prove two policies equal.

\section{Conclusion}\label{sec:conclusion}

At fine thresholds, geometric estimation with one more reference is significantly more accurate than all eight learned
models on two references on Cambridge and NAVER B1 and never significantly worse; with the same three references, the more accurate approach depends on the scene. The gain from the learned
translation magnitude depends on reference selection. References from other locations cause failures at separable stages: aligning on the nearby references largely restores VGGT, Pi3X
and MapAnything but not Depth Anything 3, and pose-conditioned Pi3X needs their poses masked and left out of its
input-pose normalization. Comparisons should fix the matcher
and references, and report each method alone and behind one shared gate.

{\small
\bibliographystyle{ieeenat_fullname}
\bibliography{references}
}
\end{document}